\documentclass[letterpaper, 10 pt]{IEEEconf}  

\IEEEoverridecommandlockouts                              

\renewcommand{\baselinestretch}{0.955}

\usepackage{amsmath}
\usepackage{amssymb}
\usepackage{tikz}

\usepackage{amsthm}
\usepackage{xspace}
\usepackage[textsize=scriptsize]{todonotes}
\usepackage{lipsum}

\usepackage{xspace}
\usepackage{array}
\usepackage{multirow}

\usepackage{colortbl}
\definecolor{cello}{HTML}{ffe6cc}
\newcommand{\colorcello}{\cellcolor{cello}}
\usepackage{graphics} 

\newtheorem{myremark}{Remark}

\newtheorem{mylem}{Lemma}
\newtheorem{mydef}{Definition}

\newtheorem{mytheorem}{Theorem}

\usepackage{color}
\usepackage{tabularray}
\definecolor{DoveGray}{rgb}{0.4,0.4,0.4}
\definecolor{Woodsmoke}{rgb}{0.094,0.101,0.105}
\newcommand{\refineCBF}{\textsc{refineCBF}\xspace}
\newcommand{\algname}{\textsc{HJ-Patch}\xspace}
\usepackage{xcolor}
\usepackage{accents}
\usepackage{centernot}
\usepackage{algorithm}
\usepackage{algorithmic}
\usepackage{subcaption} 
\usepackage{mathtools}
\usepackage[font=small,labelfont=bf]{caption}

\newcommand{\XX}{\mathcal{X}}
\newcommand{\lie}{L_f}
\newcommand{\lieopt}{L^*_f}
\newcommand{\UU}{\mathcal{U}}

\newcommand{\R}{\mathbb{R}}

\newcommand{\zerosup}[1]{{\mathcal{\MakeUppercase{#1}}}}

\newcommand{\localset}{\mathcal{H}}
\newcommand{\localvf}{h}
\newcommand{\iter}[2]{{#1^{(#2)}}}
\newcommand{\converged}[1]{{#1}^*}
\newcommand{\boundarycells}{\partial^\zeta\localset}
\newcommand{\continuousboundarycells}{\partial\localset}
\newcommand{\activeset}{Q}
\newcommand{\neighbors}[1]{X^{#1}}

\newcommand{\viability}[1]{{\mathcal{S}(#1)}}

\newcommand{\oraclecells}{C}
\newcommand{\initialvf}{\localvf^0}
\definecolor{mygray}{gray}{0.45}

\newif\ifmargincomments
\margincommentsfalse
\ifmargincomments
\newcommand{\stmargin}[2]{{\color{cyan}#1}\marginpar{\color{cyan}\raggedright\footnotesize [ST]:#2}}

\else
\newcommand{\msmargin}[2]{}
\newcommand{\stmargin}[2]{}
\fi

\newif\ifsuggestions
\suggestionsfalse

\ifsuggestions
\newcommand\stnote[1]{\textcolor{cyan}{[ST: #1]}}
\newcommand\shnote[1]{\textcolor{violet}{[SH: #1]}}

\else
\newcommand\stnote[1]{}
\newcommand\shnote[1]{}
\fi

\makeatletter
\let\NAT@parse\undefined
\makeatother
\usepackage{hyperref}

\title{\LARGE \bf
Patching Approximately Safe Value Functions Leveraging Local Hamilton-Jacobi Reachability Analysis
}

\author{Sander Tonkens, Alex Toofanian, Zhizhen Qin, Sicun Gao, and Sylvia Herbert
\thanks{The authors are with University of California, San Diego \{\href{mailto:sander@ucsd.edu}{sander}, \href{mailto:atoofanian@ucsd.edu}{atoofanian}, \href{mailto:zhq005@ucsd.edu}{zhq005},  \href{mailto:sicung@ucsd.edu}{sicung}, \href{mailto:sherbert@ucsd.edu}{sherbert}\}@ucsd.edu.}}

\begin{document}

\maketitle

\begin{abstract}
Safe value functions, such as control barrier functions, characterize a safe set and synthesize a safety filter, overriding unsafe actions, for a dynamic system. 
While function approximators like neural networks can synthesize approximately safe value functions, they typically lack formal guarantees. 
In this paper, we propose a local dynamic programming-based approach to ``patch'' approximately safe value functions to obtain a safe value function. 
This algorithm, \algname, produces a novel value function that provides formal safety guarantees, yet retains the global structure of the initial value function. 
\algname modifies an approximately safe value function at states that are both (i) near the safety boundary and (ii)~may violate safety. 
We iteratively update both this set of ``active'' states and the value function until convergence. 
This approach bridges the gap between value function approximation methods and formal safety through Hamilton-Jacobi (HJ) reachability, offering a framework for integrating various safety methods.  
We provide simulation results on analytic and learned examples, demonstrating \algname reduces the computational complexity by 2 orders of magnitude with respect to standard HJ reachability. Additionally, we demonstrate the perils of using approximately safe value functions directly and showcase improved safety using \algname.
\end{abstract}
\thispagestyle{empty}
\pagestyle{empty}

\section{Introduction}\label{sec:intro}
The past decade has witnessed significant advances in robotics, particularly in the areas of manipulation~\cite{KroemerNiekumEtAl2021} and locomotion~\cite{MikiLeeEtAl2022}, within increasingly unstructured environments. 
These advancements have been propelled by improvements in simulation engines and machine learning techniques~\cite{LiuNegrut2021}, which allow for the implicit representation of complex interactions between robots and their surroundings. 
However, the progression in safety-critical domains, such as autonomous driving, has been constrained by the necessity for quantitative safety assurances~\cite{GarciaFernandez2015}.  
In these domains, reliance on empirical safety measures alone can lead to high-impact incidents involving humans. 
Addressing this challenge, particularly when integrating learning-based components into the autonomy stack, is a formidable task~\cite{BrunkeGreeffEtAl2021}. 
In response, safety filters, which are independent modules that can be integrated into any autonomy stack, have emerged as a popular solution.  
One prevalent approach involves the use of safe value functions defined over the state space, such as control barrier functions (CBFs), to encode and enforce safety.  
The 0-superlevel set of these value functions encodes the safe set for the system, and the spatial gradients of the value function can be used to minimally modify a safety-agnostic or otherwise unsafe control policy pointwise, using an optimization-based safety filter, to ensure a trajectory remains safe~\cite{AmesGrizzleEtAl2014, HsuHuEtAl2024}.
\begin{figure}
\vspace{0.2cm}
\centering
  \includegraphics[width=\columnwidth]{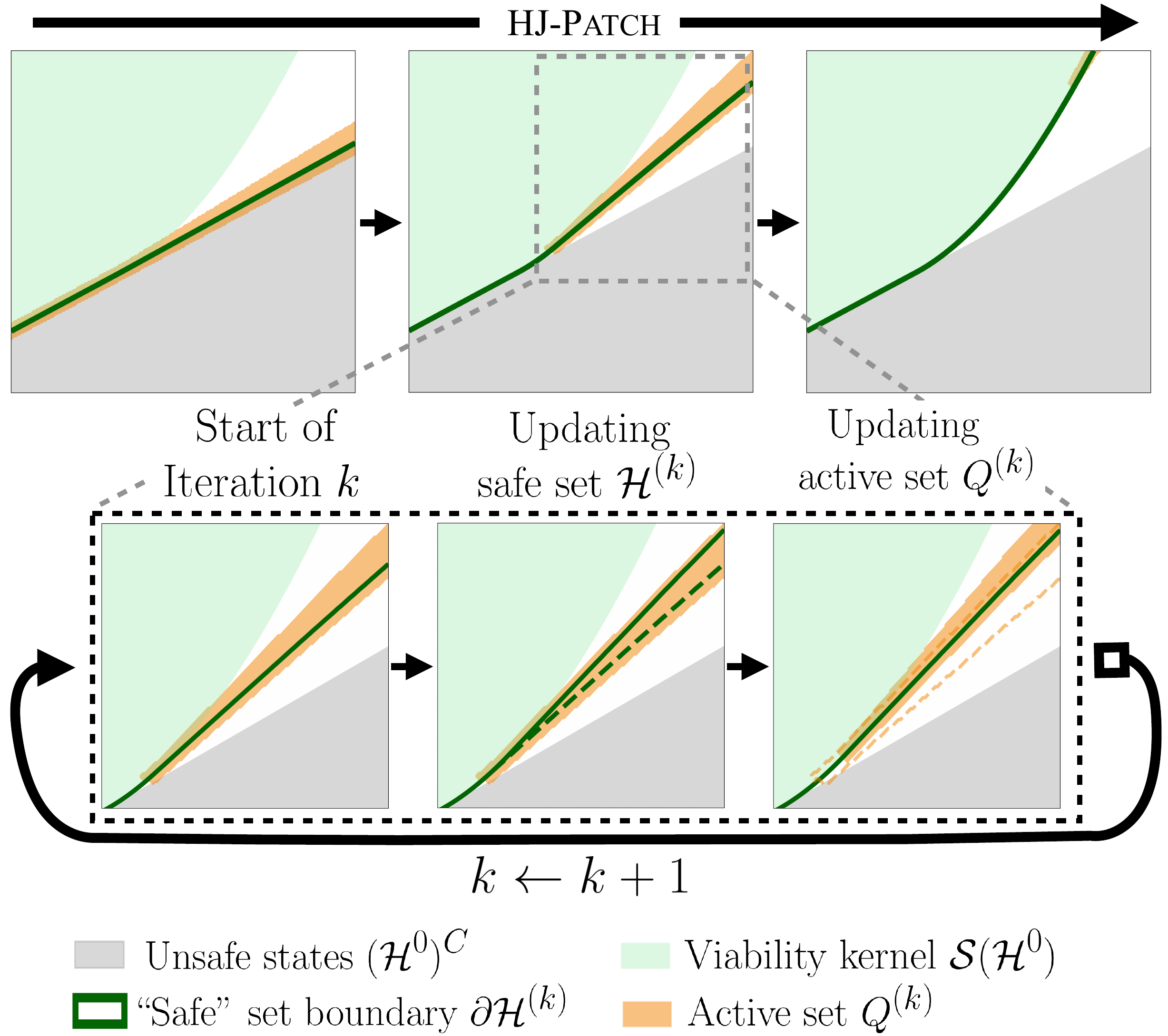}
  \caption{The process of \algname (Alg.~\ref{alg:HJR_boundary_march}) is shown in the top row from left to right. The 0-level set of an approximately safe value function $\initialvf(x)$ is plotted in the top left panel (green line), where the states such that $\initialvf(x)<0$ (gray) comprise the initial set of unsafe states. This function incorrectly classifies an unsafe region of the state space (white) as safe. 
  The initial active set for local dynamic programming $\iter{\activeset}{0}$ is shown in orange, and represents the potentially unsafe boundary states. 
  For a given iteration $k$ (below, zoomed), we compute the updated value function and plot its 0-level set (center panel) for the states in the active set. Next, we update the active set (right) for the next iteration. Upon convergence, the 0-superlevel set of $\converged{\localvf}(x)$ 
  (shaded green) is the viability kernel of the initial value function $\initialvf(x)$. \label{fig:conceptual_image}
  }
\vspace{-.5cm}
\end{figure}

For many realistic autonomy settings, suitable (safe, yet not too conservative) value functions cannot be derived analytically. 
Popular numerical methods for synthesizing value functions include solving a partial differential equation (PDE) through spatial discretization leveraging Hamilton-Jacobi (HJ) reachability~\cite{ChoiLeeEtAl2021} and sum-of-squares techniques~\cite{AhmadiMajumdar2016}. 
These both scale poorly with the dimension of the system. 
A popular recent line of work proposes approximating value functions with generic function approximation methods, most notably neural networks~\cite{DawsonGaoEtAl2023, LindemannRobeyEtAl2021, BansalTomlin2021}. 
However, these typically do not have formal guarantees.

Finding a safe value function for a nonlinear system can be posed as an optimization problem over an infinite function space. 
These problems are then solved such that the conditions hold (or are penalized) at a set of sampled points~\cite{DawsonQinEtAl2021} from, e.g. expert demonstrations. 
Then, for a restricted class of functions, it is possible to leverage Lipschitz constant estimation techniques to guarantee that satisfying stronger versions of safety, e.g. CBF, inequalities at a dense set of sampled points translates to a locally safe value function~\cite{RobeyHuEtAl2020}. 
However, this requires very fine, typically uniform, sampling of the state space. 
Additionally, these guarantees do not hold for neural network parameterized value functions.

Another line of work approximates CBFs or other safety encoding value functions by assuming unbounded control and a fully actuated system, e.g. directly using a signed distance function as CBF~\cite{LongQianEtAl2021}. However, like for learned value functions, using a signed distance function as CBF does not guarantee safety. We refer to all such approximative methods as approximately safe value functions.

In this work, we propose a technique that can verify and update approximately safe value functions (acquired by, e.g. learning-based approaches, hand-designed methods, or through invalid assumptions) to obtain strict safety assurances.
Specifically, we selectively and locally refine this initial value function by spatially discretized \textit{local} dynamic programming (DP), through HJ reachability analysis. 
Critically, in comparison to our prior work \refineCBF~\cite{TonkensHerbert2022}, we only update potentially unsafe regions around the value function's safe set boundary. 
By only updating local regions of the state space, the scalability of our DP-based approach is strongly improved compared to global updating. 
Upon convergence of our proposed iterative algorithm, we obtain a value function that encodes and enforces safety, along with a corresponding control invariant safe set.

\textbf{Contributions:}
This work proposes locally modifying an approximately safe value function to obtain a safe value function. 
We present our approach \algname (see Fig.~\ref{fig:conceptual_image} for a visual depiction), and prove that, given any initial value function, \algname is guaranteed to find a value function whose 0-superlevel set is the largest control invariant (and therefore guaranteed safe) subset of the initial value function's 0-superlevel set with a very limited (comparatively to its global counterpart) number of total queries of the recursive DP-based reachability computation. 
The associated value function both encodes and enforces (through a minimally modifying filter) safety. 
Our simulation experiments demonstrate $(i)$ the necessity of refining approximate value functions and $(ii)$ the computational benefit (yielding $10-100$ times speedup) of only selectively updating potentially unsafe states around the boundary of the learned value function. 
We demonstrate the computational benefits of \algname and empirically validate the aforementioned safety guarantees on the 2-dimensional adaptive cruise control problem. 
Next, we highlight \algname's advantages when updating learned almost CBFs, on two quadcopter models, and we evaluate safety under adversarial nominal controllers; highlighting the perils of directly using approximately safe value functions.

\textbf{Organization:}
We proceed with a brief overview of safe value functions, including control barrier functions and HJ reachability-based value functions in Section~\ref{sec:prelims}. 
Next, we present \algname, an algorithm for minimally updating approximately safe value functions (Section.~\ref{sec:patching_overview}) and prove the associated theoretical guarantees (Section~\ref{subsec:theory_guarantees}). 
We provide simulation experiments in Section~\ref{sec:results} and conclude with the merits of our approach and avenues of future work in Section~\ref{sec:conclusions}.

\section{Preliminaries}\label{sec:prelims}
We consider a dynamical system
\begin{equation}\label{eq:dynamics}
    \dot x = f(x,u), \quad s\in[t,t'],
\end{equation} 
with time $s$, state $x \in \XX \subseteq \R^n$, and input $u \in \UU \subseteq \R^m$, where $\UU$ is a compact set. 
Assume $f: \XX \times \UU \mapsto \R^n$ is uniformly continuous, bounded, and Lipschitz continuous in~$x$. 
Additionally,  assume the control signal $u(\cdot)$ is drawn from measurable control functions $u(\cdot) \in \mathbb{U} := \{ u: [t,t'] \mapsto \UU, u (\cdot) \text{ is measurable}\} $. 
Under these assumptions there exists a unique trajectory $\xi_{x, t}^{u(\cdot)}(s)$ 
starting at $x, t$ under control signal $u(\cdot)$. 

\begin{mydef}[Control invariant set]\label{def:CIset}
    A set $\mathcal{C}$ is control invariant if for any state $x \in \mathcal{C}$, there exists $u(\cdot) \in \mathbb{U}$ such that the trajectory $\xi_{x,t}^{u(\cdot)}(s) \in \mathcal{C}$ for all $s\in[t, \infty)$.
    \end{mydef}
    
    Note that in this paper we consider the empty set $\emptyset$ to be control invariant.

\begin{mydef}[Viability kernel,~\cite{AubinBayenEtAl2011}]\label{def:viability} 
For any set $\mathcal{K}$, let the viability kernel $\viability{\mathcal{K}}$ be such that $\viability{\mathcal{K}} \subseteq \mathcal{K}$ and if there exists $\mathcal{C} \subseteq \mathcal{K}$ such that $\mathcal{C}$ is control invariant, then $\mathcal{C} \subseteq \viability{\mathcal{K}}$, hence $\viability{\mathcal{K}}$ is the largest control invariant set in $\mathcal{K}$.
\end{mydef}

\subsection{Control Barrier Functions}\label{sec:cbf}
Let $\mathcal{L}\subseteq \XX$ represent the set of failure states of the system~\eqref{eq:dynamics}. 
An element $x\in\XX$ is considered instantaneously \textit{safe} if $x \in \mathcal{L}^C$, the complement of the failure set $\mathcal{L}$.

Let a value function $\localvf: \XX \mapsto \R$ be Lipschitz continuous and let $\localset:=\{x \mid h(x) \geq 0\}\subseteq\XX$ be the 0-superlevel set of $\localset$. 
We define the Lie derivative $\lie\localvf(x):=\langle \frac{\partial h}{\partial x}, f(x, u)\rangle$ and the Hamiltonian $\lieopt \localvf(x):=\sup_{u\in\UU}\lie \localvf(x)$.

Nagumo's theorem~\cite{Blanchini1999}, provides a condition for control invariance: assume $\frac{\partial h}{\partial x}\neq 0$ for all $x \in \continuousboundarycells:=\{x \mid h(x) = 0\}$, then $\localset$ is control invariant if and only if
\begin{equation}\label{eq:nagumo}
    \lieopt \localvf(x) \geq 0 \text{ for all } x \in \continuousboundarycells.
\end{equation}

In this paper, a safe value function is considered to be any function $h$ that satisfies Nagumo's theorem, and whose 0-superlevel set does not intersect with $\mathcal{L}$. Nagumo's theorem has been extended to verify safety beyond the boundary of the safe set using Lyapunov-like functions:

\begin{mydef}[Control Barrier Function~\cite{AmesGrizzleEtAl2014}]\label{def:cbf}
Let $\localset$ denote a 0-superlevel set of a continuously differentiable value function $h: \XX \mapsto \R$, then $\localvf(x)$ is a control barrier function for \eqref{eq:dynamics} if there exists an extended class $\mathcal{K}$ function $\alpha$ such that
\begin{equation}\label{eq:cbf}
    \lieopt \localvf(x) \geq -\alpha(\localvf(x)).
\end{equation}
\end{mydef}
A control barrier function (CBF) defines a control invariant set $\zerosup{h}$~\cite{AmesCooganEtAl2019}. 
Furthermore, any choice of control law $u \in \UU$ which satisfies~\eqref{eq:cbf} will cause~\eqref{eq:dynamics} to remain in $\localset$ indefinitely.

CBFs are typically defined with respect to a given failure set $\mathcal{L}$, e.g., a set of obstacles. 
Online, at each time step, a nominal safety-agnostic policy $\hat{\pi}$ is passed through a filter that solves the following optimization problem:
\begin{equation}\label{eq:online-cbf}
    \begin{split}
        u^*(x)= \> \arg \min \limits_{u} \quad &\lVert u - \hat \pi(x) \rVert_2^2 \\
        \text{subject to}\quad & \dot{\localvf}(x) + \alpha(\localvf(x)) \ \geq 0 \\
        & u \in \UU.
    \end{split}
\end{equation}
In this paper we assume $\alpha(x)=\gamma x$, for $\gamma >0$. 
The obtained control $u^*$ is guaranteed to retain safety while minimally modifying the nominal policy. 
For control-affine dynamics,~\eqref{eq:online-cbf} is a quadratic program (QP), which can be solved in real time for, e.g., robotics applications.

For most systems, synthesizing a safe value function is hard. However, it is often possible to synthesize (learning-based or otherwise derived) a value function $\localvf(x)$ whose 0-superlevel set does not intersect with the failure set $\mathcal{L}$, but nevertheless does not satisfy Nagumo's theorem \eqref{eq:nagumo}. One such example is an almost-barrier function~\cite{QinWengEtAl2022}, which allows for violating the derivative constraint on a subset of the state space, but only leads to trajectories being temporarily non-decreasing or unsafe. 
Essentially, these methods relax~\eqref{eq:nagumo} to not hold for all states on the safety boundary, but do provide point-wise satisfaction of~\eqref{eq:nagumo} at most states.

\subsection{Hamilton-Jacobi reachability}\label{sec:hj_reachability}
Assume we are given a value function $\initialvf(x)$ and its associated 0-superlevel set $\localset^0$.
In particular, $\initialvf(x)$ characterizes the ``margin'' to failure and $\localset^0$ denotes the set of non-failure states, i.e. $\localset^0=\mathcal{L}^C$. 
We defie the associated HJ reachability value function:
\begin{equation*}
    \localvf(x,t)=\max_{u \in \UU_{[t,0]}} \min_{s \in [t, 0]} \initialvf(\xi_{x,t}^u(s)).
\end{equation*}
This value function is a control barrier value function (CBVF), see~\cite{ChoiLeeEtAl2021}, for the special case $\gamma=0$. 
A CBVF is a safe value function. 
Additionally, note that a CBVF that is continuously differentiable is a CBF. 
Given an initial state $x$ and time $t<0$, this value function outputs a scalar which, if positive, indicates that trajectories from this initial condition under the optimal control $\xi_{x, t}^{u*(\cdot)}(s)$ will maintain positive value $\localvf^0(x)$ over the time horizon $[t, 0]$, and therefore will maintain safety. 
The associated PDE for $h:\R^n \times (-\infty, 0] \mapsto \R$ can be written as:
\begin{equation} \label{eq:dpcont}
\begin{split}
    \partial_t \localvf(x,t) + \min\left[0, \lieopt{\localvf}(x,t)\right]&=0\\
    \localvf(x, 0)&=\localvf^0(x),   
\end{split}
\end{equation}
see~\cite{MitchellBayenEtAl2005}.
By definition, as $t \to -\infty$, the 0-superlevel set of:
\begin{equation}
    \converged{\localvf}(x)=\lim_{t\to-\infty} \localvf(x,t),
\end{equation}
characterizes the viability kernel of $\localset^0$, i.e. $\viability{\localset^0}=\converged{\localset}$.
In this work,  we are only interested in persistent safety and hence drop the explicit time dependence of $\localvf(x)$. 

In practice, HJ reachability is solved by spatially discretizing the state space and solving for the value function $\localvf(x)$ recursively using dynamic programming on the grid:
\begin{equation}\label{eq:dp}
\iter{\localvf}{k+1}(x) = \iter{\localvf}{k}(x) + \iter{\Delta}{k} \min\{0, \lieopt \iter{\localvf}{k}(x)\},
\end{equation}
with the superscript $(k)$ denoting the iteration of the algorithm ($\iter{\localvf}{0}(x)=\initialvf(x)$) and $\iter{\Delta}{k}$ denoting a time-step which is dynamically updated based on the magnitude of the Hamiltonians  $\lieopt\iter{\localvf}{k}(x)$, see~\cite{Mitchell2008} for details. 

As the value function $\localvf(x)$ is defined on a grid, the Hamiltonian $\lieopt\iter{\localvf}{k}(x)$ is approximated using finite difference methods, namely through (weighted) essentially non-oscillatory ((W)ENO) schemes~\cite{Shu1998} as follows:
\begin{equation}\label{eq:finite_difference}
    \lie \localvf(x) := \langle \nabla \localvf(x), f(x,u)\rangle \underbrace{=}_{\text{DP approx.}} \langle \delta^p_{\Delta}(\localvf), f(x,u)\rangle,
\end{equation}
with $\delta^p_{\Delta}$ a finite difference operator $p$, i.e. the operator includes $p$ neighbors in every dimension, i.e. $\delta^p_{\Delta}(\localvf[j])$, the value of $\localvf(x)$ at cell $j$, $\localvf[j]$, depends on terms $\localvf[j-p], \cdots, \localvf[j+p]$ in every dimension independently. 
Simultaneously, we define the order $p$ neighboring set of a state $x$ as

\begin{equation}
    X^p = \{y \in \XX \mid \sum_{i=1}^n \lvert y_i - x_i\rvert \leq p\}.
\end{equation}

\section{Patching approximately safe value functions}\label{sec:patching}
We present $\algname$ (Algorithm~\ref{alg:HJR_boundary_march}, visually depicted in Fig.~\ref{fig:conceptual_image}), an efficient algorithm to obtain the viability kernel of the 0-superlevel set of a value function by minimally and locally updating its spatially-discretized value function iteratively. 
Our algorithm is particularly efficient when updating an initial value function that satisfies~\eqref{eq:nagumo} at most states, i.e. an approximately safe value function, in which often only a small subset of states require updating. 
As highlighted in Section~\ref{sec:results}, for reasonable initial value functions, our proposed algorithm provides significant speedup, improving the scalability of formal safety analysis leveraging HJ reachability by up to $2$ orders of magnitude. 

We first provide some definitions which are used in Alg.~\ref{alg:HJR_boundary_march} and to provide theoretical guarantees. We fix a threshold $\zeta~>~0$ and define the states on the boundary of a set $\localset$ as
\begin{equation}\label{eq:boundary}
    \boundarycells=\left\{x  \>: \> \lvert \localvf(x) \rvert \leq \zeta\right\}.
\end{equation}

Inspired by Nagumo's Theorem~\eqref{eq:nagumo}, we next define the notion of a quasi-control invariant set for a discretized state space.
\begin{mydef}[Quasi-control invariant set $\localset$]\label{def:quasiCI}
Let $\localvf(x)$ be defined on a discretized state-space and $\localset$ its 0-superlevel set. Assuming $\frac{\partial \localvf}{\partial x}\neq 0$ for all $x \in \boundarycells$, then $\localset$ is quasi-control invariant if and only if
\begin{equation}
    \lieopt\localvf(x) \geq 0 \text{ for all } x \in \boundarycells.
\end{equation}
\end{mydef}

\subsection{Algorithm overview}\label{sec:patching_overview}

Leveraging the structure of infinite-time HJ reachability problems, the 0-superlevel set of the value function obtained with our proposed approximate dynamic programming algorithm will correspond to the global HJ reachability's value function's 0-superlevel set, see Theorem~\ref{thm:algsafe} for details.
 
The required input to our algorithm is a value function $\initialvf(x)$. 
The states near the ``safe'' boundary (line~\ref{line:init_active}) form the set $\iter{\activeset}{0}$ of initial active states to be updated, $\iter{\activeset}{0}=\boundarycells^0$. 

If we have access to an oracle that guarantees that there is a subset of the state space $C\subseteq \XX$ where $\lieopt{\initialvf}(x)\geq0$, such as in~\cite{QinWengEtAl2022}, we can limit the size of the initial active set. 
This oracle can be obtained from, e.g., domain knowledge or neural network verification techniques~\cite{LiuArnonEtAl2021,FazlyabRobeyEtAl2019}. 
Any oracle-certified states $x\in C$ are removed from the initial active set (line~\ref{line:remove_oracle}). 
Then, at each iteration, the value function $\iter{\localvf}{k}(x)$ is updated at the active states $\iter{\activeset}{k}$ using the discrete-time HJ update formula, Eq.~\eqref{eq:dp} (line~\ref{line:update_eq}). 
A new set of interest is formed, $\iter{R}{k}$, which is composed of the subset of states of $\iter{\activeset}{k}$ whose safety value decreased (line~\ref{line:intermediate_set}). 

\begin{algorithm}[H]
\caption{\algname}
\label{alg:HJR_boundary_march}
\begin{algorithmic}[1]
\REQUIRE $\initialvf(x):\XX \mapsto \R:$ Initial value function \\
~~~~~~~$\oraclecells\subseteq \XX:$ Oracle-certified safe cells (Optional)\\
\emph{\color{mygray}// Active set is set of potentially unsafe states}
\STATE $\iter{\activeset}{0} \gets \iter{\boundarycells}{0}\label{line:init_active}
\hspace{1.05cm}\color{mygray}\vartriangleleft$ {\color{mygray}\emph{Initialize active set}}
\STATE $\iter{\activeset}{0} \gets \iter{\activeset}{0} \setminus C $ ~~~~~~~~$\color{mygray}\vartriangleleft$~{\color{mygray}\emph{Remove oracle-certified cells}}\label{line:remove_oracle}
\STATE $k \gets 0$

\emph{\color{mygray}// Boundary is certified once $\iter{\activeset}{k}$ is empty}
\WHILE{$\iter{\activeset}{k}$ is not empty}
    \FOR{$x \in \iter{\activeset}{k}$}\label{line:for_loop}
        \STATE $\iter{\localvf}{k+1}(x) \gets \iter{\localvf}{k}(x) + \iter{\Delta}{k} \min\{0, \lieopt \iter{\localvf}{k}(x)\}$\label{line:update_eq}
    \ENDFOR \\
    \emph{\color{mygray}// Value-decreasing states form set of interest}
    \STATE $\iter{\localset}{k+1} \gets \{x \mid \iter{\localvf}{k+1}(x) \geq 0\}$ 
    \STATE $\iter{R}{k+1} \gets \{x \mid \iter{\localvf}{k+1}(x) \neq \iter{\localvf}{k}(x) \wedge x \in \iter{\activeset}{k}$ \}\label{line:intermediate_set}\\
    \emph{\color{mygray}// Pad set with neighbors if near safe set boundary}
    \STATE $\iter{\activeset}{k+1} \gets \mathbf{pad}^p(\iter{R}{k+1}) \cap \iter{\boundarycells}{k+1}$\label{line:active_set_update} 
    \STATE $k \gets k+1$
\ENDWHILE
\RETURN $\converged{\localvf}(x)=\iter{\localvf}{k}(x)$~ ~$\color{mygray}\vartriangleleft$\emph{\color{mygray} ~Converged value function}\label{line:converged}
\end{algorithmic}
\end{algorithm}

\noindent Next (line~\ref{line:active_set_update}), this set is augmented by its neighbors, where $\mathbf{pad}^p$ denotes padding a space-discretized set with its $p$ neighbors in every dimension (similar to the Minkowski sum with a ball of small radius in continuous space):
\begin{equation}\label{eq:padding}
    \mathbf{pad}^p(R) = \bigcup_{r \in R} \neighbors{p}(r).
\end{equation}

This procedure repeats (line~\ref{line:for_loop}) until the active set is empty, at which point the algorithm has converged. 
The algorithm returns the converged value function $\converged{\localvf}(x)$ (line~\ref{line:converged}). 

\subsection{Theoretical guarantees}\label{subsec:theory_guarantees}
We first prove that Algorithm~\ref{alg:HJR_boundary_march} recovers a control invariant set if it converges (Lem.~\ref{lem:CI_set}); Next we show that its associated safe set is a superset of the viability kernel (Lem.~\ref{lem:superset}) at every iteration. 
Together, this proves that Alg.~\ref{alg:HJR_boundary_march} outputs the viability kernel if it converges (Thm.~\ref{thm:algsafe}). 
As Alg.~\ref{alg:HJR_boundary_march} is built upon DP-based methods, we note that the obtained results hold only for a discretized state space. 
We provide a further discussion in Section~\ref{sec:discussion_theory}.

\begin{mylem}[Algorithm~\ref{alg:HJR_boundary_march} converges to a control-invariant set]\label{lem:CI_set}
    If Alg.~\ref{alg:HJR_boundary_march} converges, then the 0-superlevel set of the value function $\converged{\localvf}(x)$, $\converged{\localset}$, obtained upon termination of the algorithm is  quasi-control invariant.
\begin{proof}
   Recall by Def.~\ref{def:quasiCI} that $\localset$ is quasi-control invariant if and only if $\lieopt{\localvf}(x)\geq 0$ for all $x\in\boundarycells$. 
   Specifically, we seek to show that for every iteration $k$, every boundary cell not in the current active set has a positive Lie derivative, i.e. if $x \in \iter{\boundarycells}{k} \setminus \iter{\activeset}{k}$, then $\lieopt\iter{\localvf}{k}(x) \geq 0$. Then, if Alg.~\ref{alg:HJR_boundary_march} converges, i.e. $\activeset = \emptyset$, we either have (i) $\boundarycells = \emptyset$ implying $\localset = \emptyset$ or (ii) $\lieopt\localvf(x) \geq 0$ for all $x \in \boundarycells$.
    
    We proceed by induction. 
    Recall that $\iter{\activeset}{0}=\iter{\boundarycells}{0} \setminus C$. 
    We have to certify that if $x \in \iter{\boundarycells}{0} \setminus \iter{\activeset}{0}=\iter{\boundarycells}{0} \cap C$, then $ \lieopt{\iter{\localvf}{0}}(x) \geq 0$, which is guaranteed for all $x\in \oraclecells$. 
    
    Next, for iteration $k$ we assume that if $x \in \iter{\boundarycells}{k} \setminus \iter{Q}{k}$, then $\lieopt \iter{\localvf}{k}(x) \geq 0$. We hence need to show that $x \in \iter{\boundarycells}{k+1} \setminus \iter{Q}{k+1} $ implies $\lieopt \iter{\localvf}{k+1}(x) \geq 0$, or its contrapositive; if $\lieopt \iter{\localvf}{k+1}(x) < 0$, then $x \in \iter{\activeset}{k+1}$ for all $x \in \iter{\boundarycells}{k+1}$. For all $x\in \iter{\boundarycells}{k+1}$ it is then sufficient to show $x\in\iter{\activeset}{k+1}$ for the following scenarios: (i) $\lieopt \iter{\localvf}{k+1}(x)=\lieopt\iter{\localvf}{k}(x) < 0$ and (ii) $\lieopt \iter{\localvf}{k+1}(x) \neq \lieopt \iter{h}{k}(x)$.

    For (i), by assumption, if $\lieopt\iter{\localvf}{k}(x)<0$, then $x \in \iter{\activeset}{k}$. As $\lieopt\iter{\localvf}{k}(x) < 0$, by~\eqref{eq:dp}, $\iter{\localvf}{k+1}(x)\neq\iter{\localvf}{k}(x)$, hence $x\in \iter{R}{k+1}$, and by extension $x \in \iter{\activeset}{k+1}$ if $x\in\iter{\boundarycells}{k+1}$.
    
    For (ii), by~\eqref{eq:finite_difference}, $\lieopt \iter{\localvf}{k+1}(x) \neq \lieopt \iter{h}{k}(x)$ implies that there exists a state $y \in \neighbors{p}$ such that $\iter{\localvf}{k+1}(y) \neq \iter{h}{k}(y)$. 
    Hence, by Alg.~\ref{alg:HJR_boundary_march} (L\ref{line:intermediate_set}), $y \in \iter{R}{k+1}$. By definition, $x$ is a neighbor of $y$ (bijective mapping), hence $x \in \mathbf{pad}^p(\iter{R}{k+1})$. Then, $x \in \iter{\activeset}{k+1}$ if $x \in \iter{\boundarycells}{k+1}$ (Alg.~\ref{alg:HJR_boundary_march} (L\ref{line:active_set_update})).
\end{proof}
\end{mylem}

The proof leverages the fact that HJ reachability is implemented using finite difference methods. 
Specifically, we have designed Alg.~\ref{alg:HJR_boundary_march} such that the active set considers all states that might have an updated Hamiltonian at the next iteration. 
We also point out the following:
\begin{myremark}
    It is possible that positive-valued states at an iteration $k$ (or oracle-safe cells at iteration $k=0$) become part of the active set at a later iteration through padding. 
    We only provide guarantees for the value function $\converged{\localvf}(x)$ upon convergence, so currently positive-valued states could become part of the unsafe set at a later iteration. 
\end{myremark}

We assert that if Alg.~\ref{alg:HJR_boundary_march} converges, the 0-superlevel set $\converged{\localset}$ (with $\iter{\localvf}{0}(x) = \initialvf(x)$) is not only quasi-control invariant, but moreover is the viability kernel $\viability{\localset^0}$ of the \textit{initial} safe set $\localset^0$. 
To show this, we first restate the following Lemma from~\cite{HerbertBansalEtAl2019}.

\begin{mylem}[{{Optimistic global warm-start HJ reachability recovers the viability kernel~\cite[Thm. 2]{HerbertBansalEtAl2019}}}]\label{lem:optimistic} 
Let $\localvf^0(x)$ be the initial function and $\converged{\localvf}(x)$ be the value obtained upon convergence of~\eqref{eq:dpcont}. Then, there exists $m:\XX\mapsto\R$ such that $\viability{\localset^0}=\{x \mid m(x)\geq0\}$, $h^0(x)\geq m(x)$ for all $x\in\XX$ and $\converged{\localvf}(x)=m(x)$ for all $x\in\XX$.
\end{mylem}
Summarizing, if HJ reachability is initialized such that it over-approximates which states are safe, the standard HJ reachability algorithm will converge to a value function $\converged{\localvf}(x)$ whose 0-superlevel set characterizes the viability kernel of the initial function $\localvf^0(x)$'s 0-superlevel set. 

\begin{mylem}\label{lem:superset}
    [Algorithm~\ref{alg:HJR_boundary_march}'s safe set upon convergence is a superset of the viability kernel] 
    If Alg.~\ref{alg:HJR_boundary_march} converges, then, initializing with $\iter{\localvf}{0}(x) = \localvf^0(x)$, we have $\converged{\localset} \supseteq \viability{\localset^0}$ upon convergence.
    
    \begin{proof}
        We denote $\Lambda$ as the single-step (i.e. within the while loop) operator for Alg.~\ref{alg:HJR_boundary_march} and $\Gamma$ as the single-step operator for standard reachability, i.e. Eq.~\eqref{eq:dp}. 
        It applies to both $\localvf(x)$ and $\localset$. 
        Hence, $\iter{\localset}{k+1}=\Lambda (\iter{\localset}{k})$. 
        Lastly, we note that $\localvf(x) \geq g(x)$ for all $x$ if and only if $\localset \supseteq \mathcal{G}$, with $\localset$ and $\mathcal{G}$ the 0-superlevel sets of $\localvf(x)$ and $g(x)$.
    
        Then, if for every iteration $k$, $\iter{\localset}{k} \supseteq \viability{\localset^0}$, we converge to a superset of $\viability{\localset^0}$. 
    
        We proceed by induction. 
        By definition of the viability kernel (Def.~\ref{def:viability}), we have $\iter{\localset}{0} \supseteq \viability{\localset^0}$. 
        
        Next, for iteration $k$ we assume $\iter{\localvf}{k}(x)$ is such that $\iter{\localset}{k} \supseteq \viability{\localset^0}$. 
        We hence need to show that $\iter{\localset}{k+1} \supseteq \viability{\localset^0}$. 
        Particularly, we will show that $\iter{\localset}{k+1} \supseteq \mathcal{G}$, and $\mathcal{G}\supseteq \viability{\localset^0}$, for $\mathcal{G}=\Gamma (\iter{\localset}{k})$.
    
        By construction, for any value function $\localvf(x)$, by Alg.~\ref{alg:HJR_boundary_march}, we have $\Lambda(\localvf(x)) \geq \Gamma(\localvf(x))$, hence $\Lambda(\localset) \supseteq \Gamma(\localset)$. 
        Thus, $\iter{\localset}{k+1} =\Lambda(\iter{\localset}{k}) \supseteq \Gamma(\iter{\localset}{k}) = \mathcal{G}$. 
        
        It remains to show that $\mathcal{G}$ is such that $\mathcal{G} \supseteq \viability{\localset^0}$. 
        By Lemma~\ref{lem:optimistic} and given $\iter{\localset}{k} \supseteq \viability{\localset^0}$, applying standard reachability~\eqref{eq:dp} recursively to $\iter{\localvf}{k}(x)$ recovers a value function $\hat{\localvf}^*(x)=\Gamma \circ \cdots \circ \Gamma(\iter{\localvf}{k}(x))$, such that $\hat{\localset}^*=\viability{\localset^0}$. 
        Noting that $\Gamma$ is a contraction mapping, we have $\viability{\localset^0} = \Gamma \circ \cdots \circ \Gamma (\iter{\localset}{k}) \subseteq \Gamma(\iter{\localset}{k}) =\mathcal{G}$. 
    
        Combining, we have $\iter{\localset}{k+1} \supseteq \viability{\localset^0} $, concluding the proof.
    \end{proof}
\end{mylem}

This proof leverages that each update of Alg.~\ref{alg:HJR_boundary_march} contracts the \textit{safe} set less than applying~\eqref{eq:dp} on the full grid, and that recursively applying~\eqref{eq:dp} recovers the viability kernel of the initial \textit{safe} set at any iteration. 

Below, we combine Lemmas~\ref{lem:CI_set} and~\ref{lem:superset} to show that Alg.~\ref{alg:HJR_boundary_march} recovers the viability kernel. 

\begin{mytheorem}[Algorithm~\ref{alg:HJR_boundary_march} recovers the viability kernel of the initial candidate safe set]\label{thm:algsafe}
    If Alg.~\ref{alg:HJR_boundary_march} converges, then, initializing with $\iter{\localvf}{0}(x)=\localvf^0(x)$, we have $\converged{\localset} = \viability{\localset^0}$. 

\begin{proof}
    Combining Lemma~\ref{lem:CI_set} and Lemma~\ref{lem:superset} and noting that the viability kernel is defined as a set's largest control invariant subset (Def.~\ref{def:viability}), we directly obtain $\converged{\localset} = \viability{\localset^0}$.
\end{proof}
\end{mytheorem}

\subsection{Discussion of theoretical results}\label{sec:discussion_theory}
A couple of remarks are in order. 
First, Theorem~\ref{thm:algsafe} holds under the assumption that Alg.~\ref{alg:HJR_boundary_march} converges.
\begin{myremark}[Convergence of Alg.~\ref{alg:HJR_boundary_march}]
    The set of value functions obtained with Alg.~\ref{alg:HJR_boundary_march}, $\{\iter{\localvf}{k}(x) \mid k \in \mathbf{N}\}$, is monotonically decreasing with $k$ (line~\ref{line:update_eq}). 
    Moreover, $\iter{\localvf}{k}(x)$ is bounded below, as states with negative value larger than $\zeta$ are excluded from the update (line~\ref{line:active_set_update}). 
    It is possible that a state enters the active set at iteration $k$, $x \in \iter{\activeset}{k}$ and $x\in\iter{\activeset}{k'}$ for all $k'>k$. 
    In this case $\lieopt \iter{\localvf}{k}(x)$ is bounded below by a negative function that approaches $0$ as $k\to \infty$.
    Hence, convergence is guaranteed in infinite-time. 
    When implemented numerically, states whose value function is only minimally updated are excluded from the safe set, so Alg.~\ref{alg:HJR_boundary_march} always converges in practice.
\end{myremark}

Next, we discuss the validity of using a state-discretized value function to encode and enforce safety for systems with a continuous state space. 
In theory the numerical approximation scheme (implementation of dynamic programming) is exact at the spatial grid states at all times $\localvf(x,t)$, including at convergence $\converged{\localvf}(x)$. 
In addition, HJ reachability and therefrom derived safety methods assume that the grid density is high enough such that in practice for any continuous boundary state $x\in\continuousboundarycells$, the linearly interpolated value of $\localvf(x)$, $\bar{\localvf}(x)$, is such that if $\lieopt \bar{\localvf}(x)\geq 0$ then $\lieopt \localvf (x) \geq 0$. 

Having a sufficiently fine spatial discretization for a specific problem is an assumption underlying any dynamic programming-based HJ reachability approach~\cite{Mitchell2008}. 
To guarantee safety in continuous space, a bounded disturbance (in each dimension individually), which is proportionally scaled to the grid density and the maximum Lipschitz constant of $\lieopt{\localvf}(x)$ over the grid, may be added when computing the value function iteratively in Eq.~\eqref{eq:dp}. 
This approach leverages ideas from safety guarantees for learned CBFs~\cite{RobeyHuEtAl2020}. 
Importantly, when this assumption is invalid (i.e. when the discretized grid is too coarse), \algname does not hold strict safety guarantees but nevertheless greatly reduces the \textit{degree} of unsafety.  
As seen in the empirical results (Sec.\ref{sec:learned_4dim}, \ref{sec:learned_6dim}), \algname provides significant improvements to safety even in high dimensions where the dense grid assumption is invalid.

Lastly, for control invariant sets, invariance by itself is not a sufficient condition to use $\converged{\localvf}(x)$ in the CBF safety filter~\eqref{eq:online-cbf}. 
Importantly,~\eqref{eq:online-cbf} limits the rate at which a trajectory is permitted to approach unsafety. There is a separate class of safety filters, see~\cite{WabersichTaylorEtAl2023}, that only require invariance, but are practically less favorable due to bang-bang behavior~\cite{BajcsyBansalEtAl2019}.
\begin{myremark}
The obtained value function $\converged{\localvf}(x)$, is a CBVF~\cite{ChoiLeeEtAl2021} for an appropriately chosen class $\mathcal{K}$ function $\alpha$ inside its 0-superlevel set, $\converged{\localset}$. 
In practice, the obtained value function can have relatively large gradients in the interior of the safe set. To normalize these gradients, one can reconstruct a signed distance function to $\converged{\localset}$ or consider a large threshold $\zeta$ in~\eqref{eq:boundary} to enlarge the size of the boundary considered for updating at each iteration in Alg.~\ref{alg:HJR_boundary_march}. 
\end{myremark}

\section{Experiments}\label{sec:results}
We evaluate the performance of \algname in simulation on 3 different problem formulations. 
First, we empirically demonstrate that the discretized-state convergence and safety guarantees of \algname shown in Thm.~\ref{thm:algsafe} translate to continuous state-space guarantees for the 2-dimensional adaptive cruise control example, see e.g.~\cite{AmesCooganEtAl2019}, for a fine spatial discretization. 
Next, we highlight that \algname is well-suited to modify learned CBFs, on a 4-dimensional quadrotor that ignores lateral motion and a 6-dimensional planar quadrotor. For these experiments, to compare to global HJ reachability (implemented with \refineCBF), we use a relatively coarse grid to retain computational tractability for \refineCBF while providing a fair comparison.

We provide an open-source implementation of our approach at \href{https://github.com/UCSD-SASLab/HJ-Patch}{https://github.com/UCSD-SASLab/HJ-Patch}, which builds wrappers around the \textsc{OptimizedDP}\xspace toolbox~\cite{BuiGiovanisEtAl2022} and the jax-based \textsc{HJReachability}~\cite{Schmerling} toolbox.

\subsection{Expert barrier for adaptive cruise control}
We showcase the effectiveness of \algname for two different initial approximately safe value functions on the adaptive cruise control example in~\cite{AmesCooganEtAl2019}. 
First, we consider initializing with the signed distance function to the obstacle. This is a common technique to design approximate CBFs for which safety guarantees do not hold for systems with control bounds. We show that this does not guarantee safety by sampling 100 trajectories starting within the initial ``safe'' set $\localset^0$. These sampled trajectories are adversarial, i.e. without a safety filter applied they would violate safety.
Under the signed distance function as an approximate CBF, $14\%$ of trajectories lead to failure. 
Applying \algname to this initialization, see Fig.~\ref{fig:conceptual_image} for a visual depiction of the iterative procedure, requires $1.97\%$ of updates compared to the global \refineCBF. The resulting empirical safety analysis leads to no failures of the 100 sampled trajectories.

\begin{figure}[!t]
\centering
\vspace{.2cm}
  \includegraphics[width=\columnwidth]{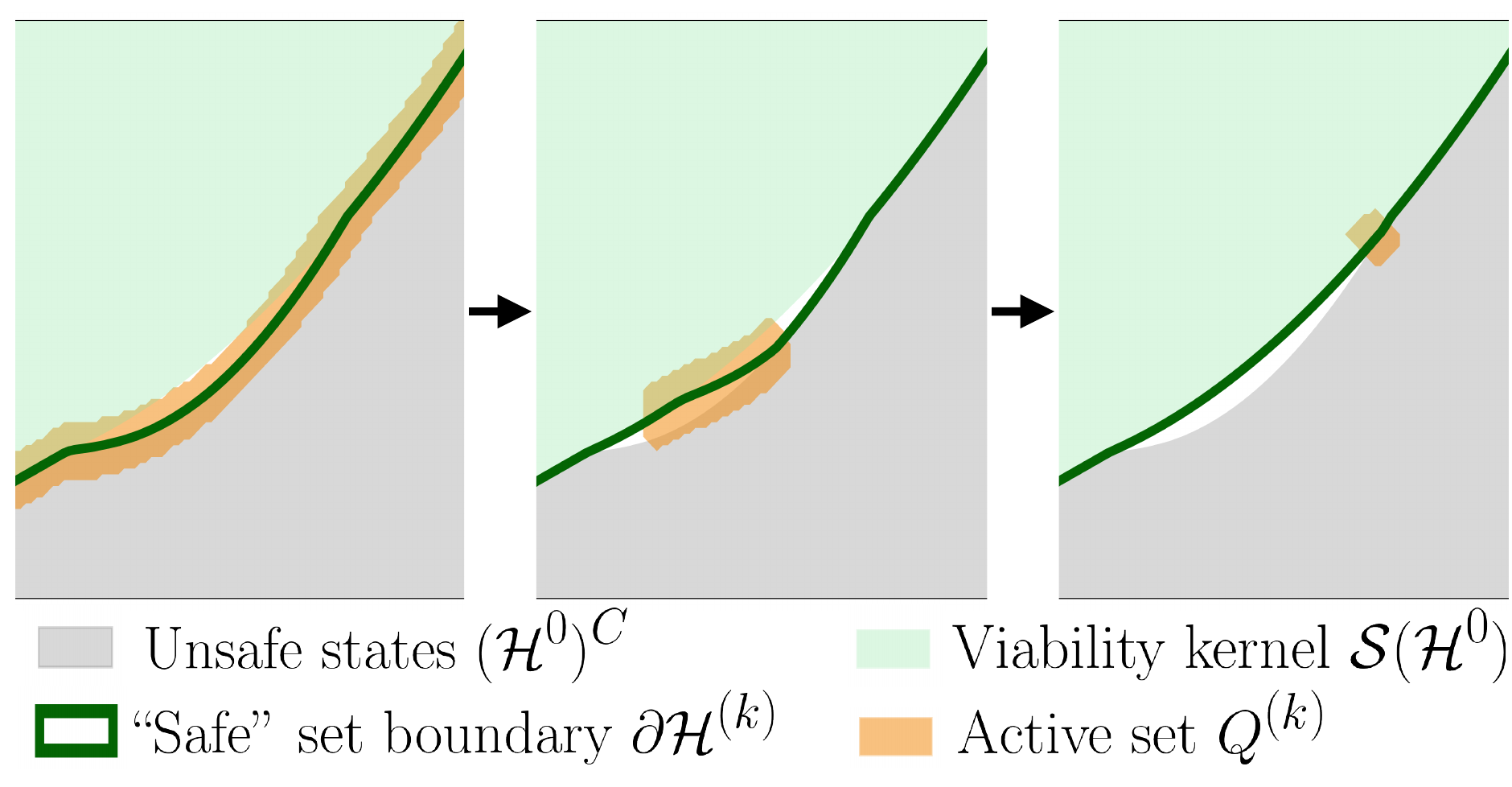}
  \caption{\footnotesize \algname iteratively updates an approximately safe value function for the adaptive cruise control example. Small errors in the initial value function result in incorrectly classifying an unsafe region of the state space (white region, right) as safe. These are efficiently ``patched'' (left to right), resulting in a safe value function at convergence (right).}
  \label{fig:figure_ACC}
  \vspace{-.5cm}
\end{figure}

For the second initialization, we artificially perturb the value function that characterizes the viability kernel (when applying \refineCBF) around the safe set boundary in a subset of the state space, to mimic a learned approximately safe value function, see Fig.~\ref{fig:figure_ACC} for details. 
The perturbed initial value function empirically fails $51\%$ of the time. 
Trajectories that pass through an area where the learned barrier is wrong might lead to immediate failure.
This highlights the perils of naively applying an approximately safe value function as a safety filter. 
Applying \algname requires updating only $0.85\%$ of the shares of states compared to using \refineCBF (both with the same termination settings, hence terminating at a different iteration). The resulting empirical safety analysis leads to $0\%$ failure. Both examples highlight (i) the computational speedups of \algname and (ii) the need for refining approximately safe value function.

\subsection{Learned barrier for vertical quadcopter (4-dimensional)}\label{sec:learned_4dim}
For both quadrotor environments, we train a control policy and corresponding neural barrier function such that the quadrotor will self-right from a random initial state. We consider vertical constraints, the floor and ceiling (for 4D and 6D), and additionally horizontal constraints, 2 walls (for 6D). 
The dynamics for the planar quadrotor models are adopted from~\cite{SinghRichardsEtAl2020}. 
We compare to global HJ reachability, implemented using \refineCBF, for both environments.

\begin{figure}[t]
\centering
  \includegraphics[width=\columnwidth]{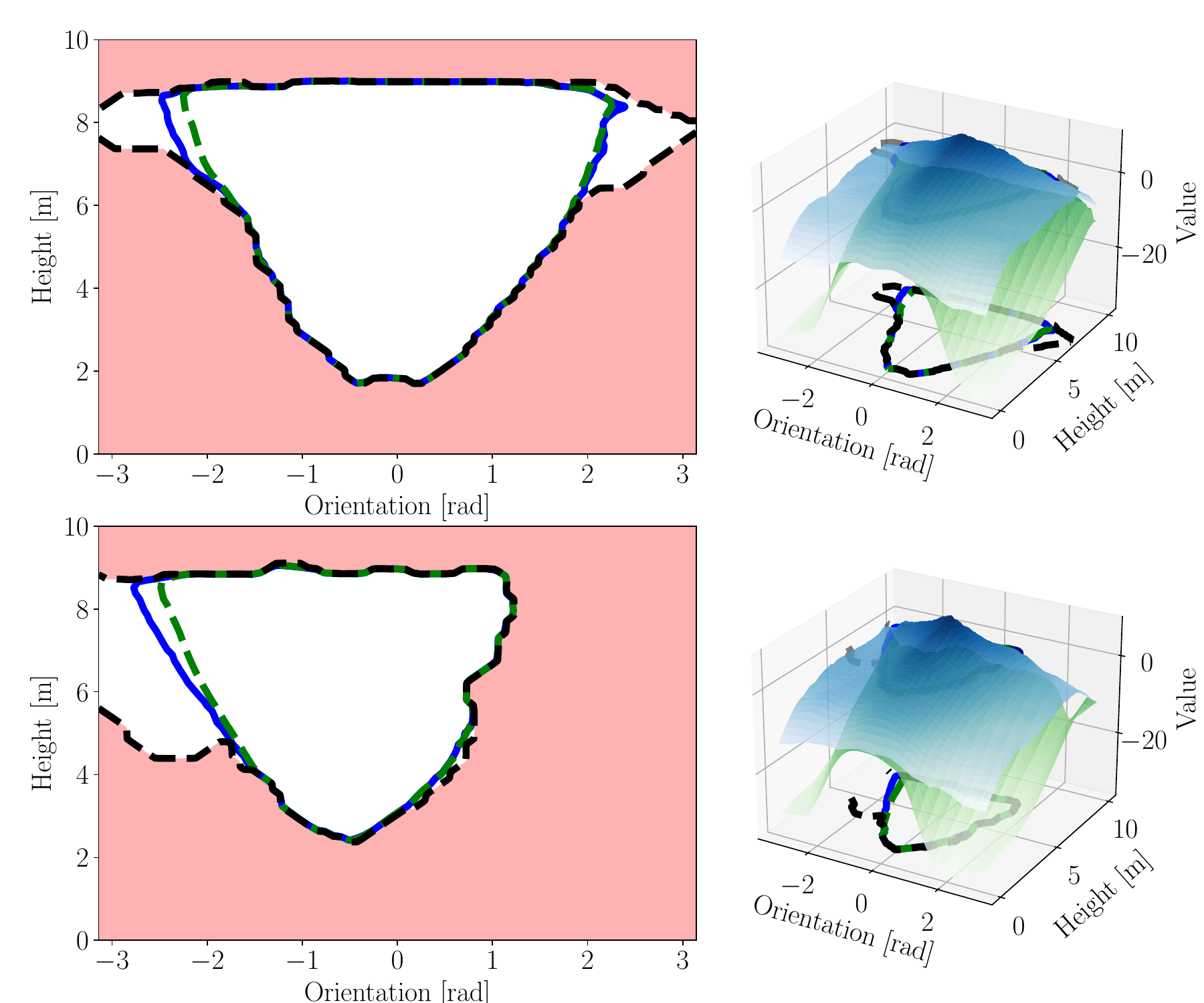}
  \caption{\footnotesize 0-superlevel set (left) and value function (right) on different 2D projections: 0 velocities (top) and large negative velocities (bottom). 
  The patched value function is in blue, the standard HJR value function is in green, and the set associated with the original neural value function is in black. 
  The patched CBVF provides a tight approximation of the global solution within and near the boundary, but has a flattened value function, hence small gradients, outside the safe set in regions that required extensive updating.}
  \label{fig:vf_4d}
\end{figure}

For the 4D quadcopter model, as seen from Fig.~\ref{fig:vf_4d}, left, for different 2D projections of the 4D viability kernel, the solution from \algname (blue) and \refineCBF (green) match closely.
Fig.~\ref{fig:vf_4d}, right shows the value function produced by \algname (blue) and \refineCBF (green). As expected, the value function from \algname closely matches the global solution near the boundary, whereas the local solution has steeper gradients near the boundary within the viability kernel, and flatter gradients outside.

\begin{figure}[!t]
\centering
  \includegraphics[width=\columnwidth]{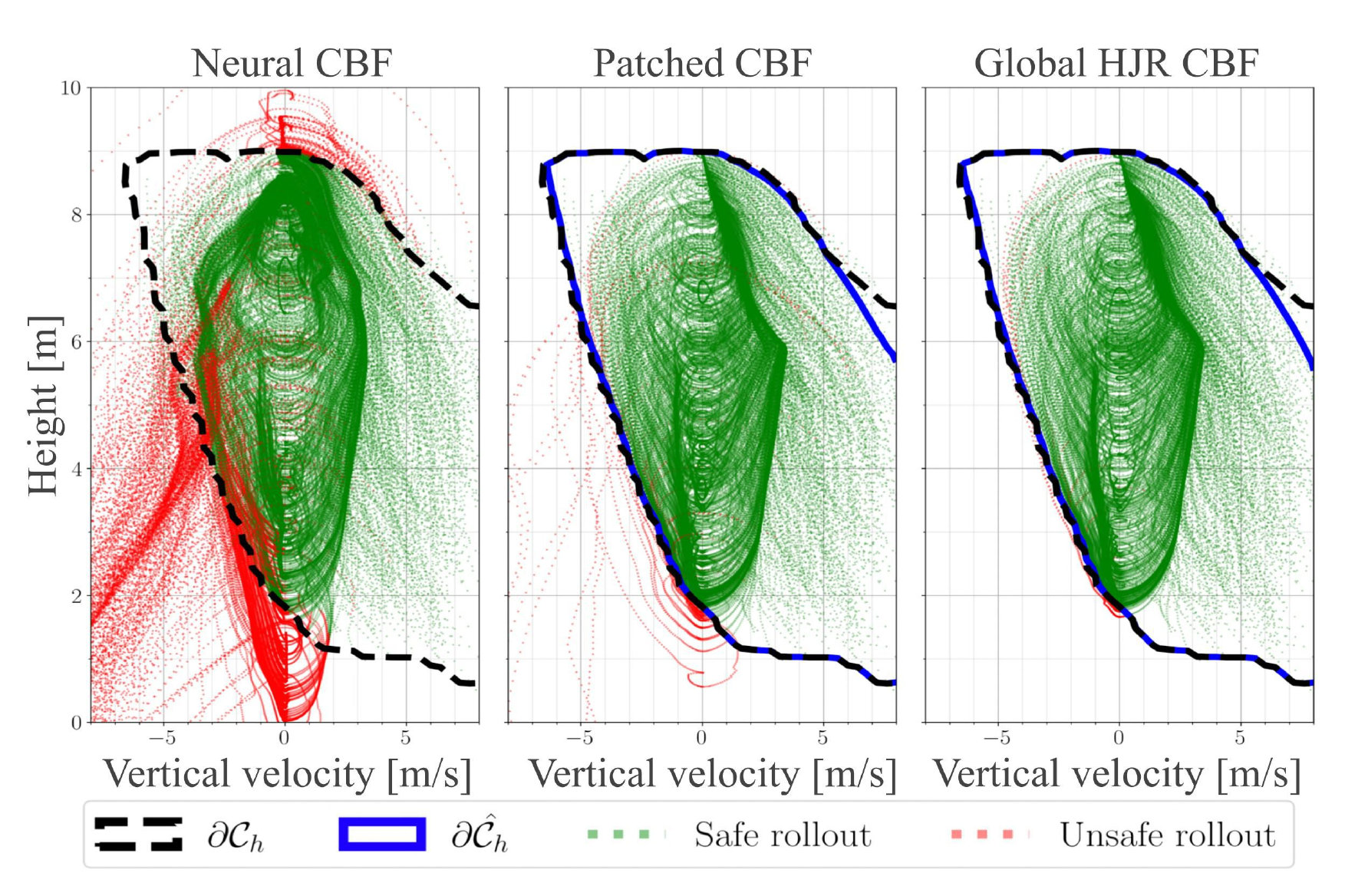}
  \caption{\footnotesize $1000$ sampled trajectories over $10s$. 
  The learned policy acts as the nominal controller, regulated with a CBF-based safety filter using the neural value function (left), the \algname value function (center), and the warm-started standard HJ value function (right). 
  The large number of unsafe trajectories (red; green is safe) for the neural value function highlights the need for patching neural barrier functions. 
  Similar behavior is observed for other learned value functions when having an adversarial nominal policy. 
  As detailed in Section~\ref{subsec:theory_guarantees}, invariance is only strictly guaranteed at the discretized states, hence resulting in non-zero unsafe trajectories for both \algname and the global HJ value function. 
  }
  \label{fig:4d_rollout}
  \vspace{-.5cm}
\end{figure}

\begin{table*}[t]
\centering

\caption{\footnotesize Quantitative comparison of \algname compared to learned barrier and global reachability for the 4D quadrotor. 
We provide a quantitative comparison of the performance of the safety filter through randomized adversarial trajectories and quantify the speedup of \algname by comparing the total number of cells that have been updated to obtain the value function. 
We observe that \algname drastically reduces the rate of unsafe trajectories compared to the learned almost CBF while being moderately more unsafe than when applying reachability on the full grid. 
However, applying reachability globally comes with a $14$x increase in computational cost.}
\label{tab:quad4d}
\centering
\setlength{\tabcolsep}{2pt}
\setlength{\extrarowheight}{1.1pt}
\begin{tabular}{l|cc|cc|cc}
 & \multicolumn{4}{c|}{Empirical safety when deployed within safety filter} & \multicolumn{2}{c}{Computational cost} \\ [1mm]
\hline
 & \multicolumn{2}{c|}{Neural nominal policy}  & \multicolumn{2}{c|}{LQR nominal policy} & \multirow{3}{*}{\begin{tabular}[c]{@{}c@{}}Hamiltonians \\computed [\#]\end{tabular}} & \multirow{3}{*}{\begin{tabular}[c]{@{}c@{}}Share of grid \\labeled safe [\%]\end{tabular}}  \\ [1mm]
 & \begin{tabular}[c]{@{}c@{}}Share of states visited \\that are unsafe [\%]\end{tabular} & \begin{tabular}[c]{@{}c@{}}Share of trajs. \\that are unsafe [\%]\end{tabular} & \begin{tabular}[c]{@{}c@{}}Share of states visited \\that are unsafe [\%]\end{tabular} & \begin{tabular}[c]{@{}c@{}}Share of trajs. \\that are unsafe [\%]\end{tabular} &     &           \\ [1mm]
\cline{2-7}
\multicolumn{1}{c|}{Learned almost CBF}      & 39.8 & 66.1  & 36.2 & 52.7  &          & 31.5      \\[1mm]
\multicolumn{1}{c|}{\refineCBF} & 0.2  & 2.0   & 0.3  & 2.4   & 9.5e7   & 21.7      \\[1mm]
\multicolumn{1}{c|}{\colorcello HJ-Patch (Ours)}         & \colorcello 1.3  & \colorcello 4.4   & \colorcello 1.3  & \colorcello 4.2   & \colorcello 7.3e6   & \colorcello 22.7     
\end{tabular}
\end{table*}

\begin{table*}
\centering
\caption{\footnotesize Quantitative comparison of \algname compared to the learned barrier and global reachability for 6D planar quadrotor. 
We provide a quantitative comparison of the performance of the safety filter through randomized adversarial trajectories and quantify the speedup of \algname by comparing the total number of cells that have been updated to obtain the safe value function. 
We observe that \algname drastically reduces the rate of unsafe trajectories compared to the learned almost CBF and has a lower unsafe rate than global reachability. 
Additionally, \algname has a $100$x lower computational cost.}
\label{tab:quad6d}
\centering
\setlength{\tabcolsep}{2pt}
\setlength{\extrarowheight}{1.1pt}
\begin{tabular}{l|cc|cc|cc}
 & \multicolumn{4}{c|}{Empirical safety when deployed within safety filter}& \multicolumn{2}{c}{Computational cost}                \\[1mm] 
\hline
 & \multicolumn{2}{c|}{Neural nominal policy}  & \multicolumn{2}{c|}{LQR nominal policy}     & \multirow{3}{*}{\begin{tabular}[c]{@{}c@{}}Hamiltonians \\computed [\#]\end{tabular}} & \multirow{3}{*}{\begin{tabular}[c]{@{}c@{}}Share of grid \\labeled safe [\%]\end{tabular}}  \\[1mm]
 & \begin{tabular}[c]{@{}c@{}}Share of states visited \\that are unsafe [\%]\end{tabular} & \begin{tabular}[c]{@{}c@{}}Share of trajs. \\that are unsafe [\%]\end{tabular} & \begin{tabular}[c]{@{}c@{}}Share of states visited \\that are unsafe [\%]\end{tabular} & \begin{tabular}[c]{@{}c@{}}Share of trajs. \\that are unsafe [\%]\end{tabular} &     &           \\ [1mm]
\cline{2-7}
\multicolumn{1}{c|}{Learned almost CBF}      & 0.4  & 2.3   & 38.6 & 57.3  &         & 4.2      \\[1mm]
\multicolumn{1}{c|}{\refineCBF} & 0.7  & 6.1   & 7.0  & 16.2   & 1.6e9   & 2.8      \\[1mm]
\multicolumn{1}{c|}{\colorcello HJ-Patch (Ours)}         & \colorcello 0.0  & \colorcello 0.5   & \colorcello 0.1  & \colorcello 0.7   & \colorcello 1.6e7   & \colorcello 3.7     
\end{tabular}
\end{table*}

In addition, we demonstrate the need for patching neural value functions by sampling trajectories ($n=1000$, $t~=~10s$) that start within the safe set and employ said value function as a CBF in a safety filter. 
This equally enables us to quantify the performance of the obtained value functions, both global and local, when used as CBVFs in a safety filter with a nominal policy that is potentially unsafe, see Tab.~\ref{tab:quad4d} for a quantitative comparison and Fig.~\ref{fig:4d_rollout} for a visual depiction. 
We select a nominal policy that results in failure for all sampled trajectories when not using a safety filter. 

Due to the coarser discretization in the higher-dimensional space, the assumption of an appropriately dense grid is violated and therefore safety is not guaranteed. However, we observe two important points: (i) despite the lack of guarantees, safety is vastly improved compared to the original learned approximately safe value function, and (ii) the share of unsafe trajectories of \algname and global HJ reachability are on the same order of magnitude, while \algname provides a $14$x computational speedup. Together these advantages make \algname useful in practice compared to naive implementation of learned approximately safe value functions.

The trajectories in Fig.~\ref{fig:4d_rollout} are visualized in the vertical position-vertical velocity plane. 
We observe that although the share of unsafe trajectories of \algname and \refineCBF (standard HJ reachability) are on the same order of magnitude, \algname's value function does not provide an exponential return to safety (leading to a comparatively larger multiplier on the share of unsafe states visited compared to the share of unsafe trajectories). 
This is likely due to the flattened gradients outside the safe set. 
Fitting a signed-distance function to the obtained value function can mitigate this behavior.

\subsection{Learned barrier for planar quadcopter (6-dimensional)}\label{sec:learned_6dim}
To provide a valid comparison to the global solution, we consider a 6D grid with relatively large spacing between grid cells to retain computational tractability. 
Again, like for the 4D vertical quadrotor, we quantitatively compare \algname to the global HJR solution and the learned almost CBF. 
We observe that the learned almost CBF requires only limited ``patching'' (see Table~\ref{tab:quad6d}, the share of cells labeled as safe upon convergence of \algname overlaps with the learned barrier for $>90$\%). 
Additionally, using the same convergence criteria, we see that \algname obtains a speedup of 2 orders of magnitude. 
In line with our expectations, the speedup of \algname becomes more significant for higher dimensional systems and for better initial approximately safe value functions. 
For the empirical safety quantification of the CBF, we observe that both global HJR and \algname improve safety, but that \algname has a lower failure rate (see Table~\ref{tab:quad6d}). 
We hypothesize that the non-zero failure rate is due to the grid not being appropriately dense.

\section{Conclusions}\label{sec:conclusions}
In this work, we propose \algname, a constructive approach to patch approximately safe value functions using \textit{local} dynamic programming. 
Our method guarantees convergence to the largest control invariant subset of the initial value function's \textit{safe} set, and the resulting safe value function can be readily used within a safety filter to maintain safety. 
As highlighted in the results, \algname enables reducing the computational complexity of HJ reachability by $2$ orders of magnitude, enabling scaling HJ reachability by $1-2$ dimensions. 
Additionally, we observe that in practice we converge to the same viability kernel or a tight over-approximation of the viability kernel that is obtained with standard HJ reachability. For a fine grid, we show empirically that we can guarantee safety. However, for coarser grids, we observe a few unsafe sampled trajectories, yet much improved compared to learned approximately safe value functions and at the same order of magnitude of global HJ reachability. 

\textbf{Future work:} We are interested in further speedups for locally updating value functions using more informed updating strategies and leveraging techniques from computational physics such as adaptive grids. 
Next, we plan to investigate how to extend the provided guarantees, i.e. recovering the viability kernel, to coarser grids for both global HJR and \algname. 
Additionally, we plan to extend our algorithm to alternatively expand to the smallest invariant superset.

\section*{Acknowledgments}
We would like to thank Dylan Hirsch for the invaluable feedback and Minh Bui for assisting in the implementation. 
\addtolength{\textheight}{-8.9cm}

\bibliographystyle{IEEEtran}
\bibliography{main, ASL_papers, SASLab}
\end{document}